\documentclass[letterpaper]{article} 
\usepackage{aaai25}  
\usepackage{times}  
\usepackage{helvet}  
\usepackage{courier}  
\usepackage[hyphens]{url}  
\usepackage{graphicx} 
\usepackage{natbib}  
\usepackage{caption} 
\usepackage{algorithm}
\usepackage{algorithmic}

\usepackage{newfloat}
\usepackage{listings}
\DeclareCaptionStyle{ruled}{labelfont=normalfont,labelsep=colon,strut=off} 
\floatstyle{ruled}
\newfloat{listing}{tb}{lst}{}
\floatname{listing}{Listing}
\usepackage{amsmath}
\usepackage{amssymb}

\usepackage{bm}
\usepackage{booktabs}
\usepackage{colortbl}
\usepackage{multirow}
\usepackage[subrefformat=parens]{subcaption}
\newif\ifshowapdx
\showapdxtrue

\newcommand{\apdxref}[2]{%
    \ifshowapdx
        \ref{#1}%
    \else
        #2%
    \fi
}

\title{Universal Post-Processing Networks for Joint Optimization of Modules\\in Task-Oriented Dialogue Systems}
\author {
    Atsumoto Ohashi,
    Ryuichiro Higashinaka
}
\affiliations {
    Graduate School of Informatics, Nagoya University, Japan\\
    ohashi.atsumoto.c0@s.mail.nagoya-u.ac.jp, higashinaka@i.nagoya-u.ac.jp
}

\usepackage{bibentry}

\begin{document}

\maketitle

\begin{abstract}
Post-processing networks (PPNs) are components that modify the outputs of arbitrary modules in task-oriented dialogue systems and are optimized using reinforcement learning (RL) to improve the overall task completion capability of the system. However, previous PPN-based approaches have been limited to handling only a subset of modules within a system, which poses a significant limitation in improving the system performance. In this study, we propose a joint optimization method for post-processing the outputs of all modules using universal post-processing networks (UniPPNs), which are language-model-based networks that can modify the outputs of arbitrary modules in a system as a sequence-transformation task. Moreover, our RL algorithm, which employs a module-level Markov decision process, enables fine-grained value and advantage estimation for each module, thereby stabilizing joint learning for post-processing the outputs of all modules. Through both simulation-based and human evaluation experiments using the MultiWOZ dataset, we demonstrated that UniPPN outperforms conventional PPNs in the task completion capability of task-oriented dialogue systems.
\end{abstract}

%
\begin{links}
    \link{Code}{https://github.com/nu-dialogue/UniPPN}
\end{links}

\section{Introduction}
Typical task-oriented dialogue systems process a single user input through multiple subtasks to produce a final response. These subtasks include (1) natural language understanding (NLU), which estimates the user's intent from the input; (2) dialogue state tracking (DST), which accumulates the user's requests up to the current turn as a dialogue state; (3) dialogue policy (policy), which determines the next action that the system should take as dialogue acts (DAs); and (4) natural language generation (NLG), which converts these DAs into a final system response. Recent research has moved beyond optimizing dedicated modules for each subtask individually. It has explored the use of reinforcement learning (RL) to train several modules based on actual dialogue experiences, thereby optimizing the overall task completion capability of the system~\citep{ni2022recent, kwan2023survey}. 

Recently, a new approach using post-processing networks (PPNs) was proposed to optimize the task completion capability of dialogue systems without directly training the modules~\citep{ohashi-higashinaka-2022-post}. In this approach, instead of training the modules directly, PPNs, which are components with trainable parameters that modify their outputs, are trained using RL. For example, \citet{ohashi-higashinaka-2022-post} implemented PPNs that post-process outputs from NLU, DST, and policy as multi-binary classification tasks using multi-layer perceptrons (referred to as BinPPN), and demonstrated that this optimization improved the overall task completion capability of the systems. In addition, a large language model (LLM)-based generative PPN (GenPPN) was proposed to post-process the natural language output from NLG, and its effectiveness has been demonstrated~\citep{ohashi-higashinaka-2023-enhancing}.

However, conventional PPN-based methods have two major limitations. First, BinPPN and GenPPN cannot be optimized jointly because of their different model architectures and training algorithms. Although it is possible to post-process the outputs of all modules by combining disjointly trained BinPPNs and GenPPNs, using multiple networks that are not jointly optimized may fail to achieve sufficient performance improvement. The second limitation is the narrow applicability of GenPPN. GenPPN aims to generate utterances that are easily understood by users and relies on a reward function that requires feedback on whether the DAs output from the policy are correctly conveyed to the user. This approach cannot be applied to an end-to-end dialogue system, such as the LLM-powered model proposed by \citet{hudecek-dusek-2023-large}, which does not explicitly output DAs, imposing significant limitations on GenPPN's applicability.

\begin{figure}
\centering
\includegraphics[width=\linewidth]{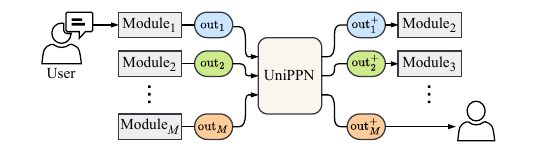}
\caption{Diagram of UniPPN. UniPPN modifies the output $\text{out}_m$ of $\text{Module}_m$ to $\text{out}_m^+$, which serves as the input for $\text{Module}_{m+1}$.}
\label{fig:unippn}
\end{figure}

In this study, we propose a universal post-processing network (UniPPN) and an optimization method that combines the strengths of two conventional PPNs. UniPPN can jointly optimize the post-processing of outputs from all modules in task-oriented dialogue systems (Figure~\ref{fig:unippn}). Our proposed method involves a single-language model-based UniPPN that post-processes the outputs from all modules as a sequence-transformation task. Additionally, we introduce a newly designed module-level Markov decision process (MDP) that extends the standard MDP paradigm and incorporates it into UniPPN's optimization algorithm. This approach enables fine-grained value and advantage estimation for each module, even with sparse feedback obtained only at the end of multi-turn dialogues, thus ensuring stable joint optimization. Because UniPPN does not require as dense feedback about DAs as GenPPN, it can be applied to a wide range of systems, including end-to-end systems that do not output DAs.

To verify the effectiveness of UniPPN across various dialogue systems, we conducted experiments using dialogue simulations based on the MultiWOZ dataset~\citep{budzianowski-etal-2018-multiwoz}. Specifically, we compared the task completion capabilities of dialogue systems equipped with conventional PPNs to those using UniPPN. The results demonstrated that UniPPN significantly outperformed conventional PPNs in terms of task completion capability. Additionally, through dialogue experiments with human users, we demonstrated that dialogue systems using UniPPN outperformed those using conventional PPNs.

\section{Related Work}
\paragraph{Modularity of Task-Oriented Dialogue Systems}
In typical pipeline task-oriented dialogue systems, dedicated modules for each subtask, such as NLU, DST, policy, and NLG, have been individually developed and optimized~\citep{zhang2020recent}. However, in recent years, methods addressing multiple subtasks using a single model have become common~\citep{ni2022recent}. For example, word-level DST~\citep{wu-etal-2019-transferable, zhao2022description} estimates the dialogue state directly from the dialogue history without requiring user intent estimation using NLU. Similarly, a word-level policy ~\citep{lubis-etal-2020-lava, wang-etal-2020-multi-domain} generates system responses directly from the dialogue state without requiring conversion from DA to system utterances by NLG. Furthermore, end-to-end dialogue systems~\citep{NEURIPS2020_e9462095, He_Dai_Zheng_Wu_Cao_Liu_Jiang_Yang_Huang_Si_Sun_Li_2022, wu-etal-2023-diacttod}, which learn all subtasks using a single model, are becoming popular. Because these end-to-end systems maintain modularity by sequentially executing each subtask, they are often referred to as modularly end-to-end systems ~\citep{qin-etal-2023-end}.

\paragraph{Online RL for Task Completion}
In response generation for task-oriented dialogue systems, it is crucial not only to maximize the probability of reference tokens in corpora but also to maximize task completion capability in actual multi-turn dialogues~\citep{kwan2023survey}. Some studies~\citep{liu-etal-2018-dialogue, tseng-etal-2021-transferable} employed online RL frameworks to train dialogue systems based on experiences obtained online from interactions with users. For example,  research has focused on optimizing DST~\citep{10.1007/978-981-99-2401-1_25} or policy ~\citep{li-etal-2020-guided, deng2024plugandplay} within pipeline systems. Additionally, \citet{zhao-eskenazi-2016-towards} demonstrated that jointly optimizing DST and policy with shared parameters outperforms systems in which the DST and policy are trained separately. Our proposed method also utilizes an online RL framework  to optimize dialogue systems. However, contrary to previous studies that focused on learning the modules, we concentrate on learning to modify the outputs of these modules.

\paragraph{Post-Processing Networks}
Methods that train modules via RL cannot be applied to dialogue systems with non-trainable modules, such as rule-based or API-based modules, as expected in real-world scenarios. To address this issue, \citet{ohashi-higashinaka-2022-post} proposed optimizing BinPPNs instead of the modules. BinPPNs modify the outputs of NLU, DST, and policy, and are optimized using online RL. Specifically, BinPPNs perform post-processing on the set of slot-value pairs output by each module through binary classification to determine whether to delete (0) or maintain (1) each pair. To handle post-processing for NLG, which outputs natural language rather than a set of slot-value pairs, \citet{ohashi-higashinaka-2023-enhancing} introduced GenPPN, which uses LLMs to paraphrase the system utterance output by NLG, to improve task completion by generating system utterances that can be easily understood by users. It is optimized through RL based on  feedback regarding whether DAs are correctly conveyed to users.

Our proposed UniPPN can process arbitrary sequences as both input and output. Therefore, contrary to BinPPN, which is limited to binary decisions of deletes/maintenance, it allows for more flexible post-processing. In addition, contrary to GenPPN, which is optimized using detailed feedback on DAs, UniPPN is optimized solely using the task success/failure signal obtained at the end of the dialogue. This makes it applicable to a wide range of systems, including word-level policies and end-to-end systems.

\begin{figure*}
\centering
\includegraphics[width=1\linewidth]{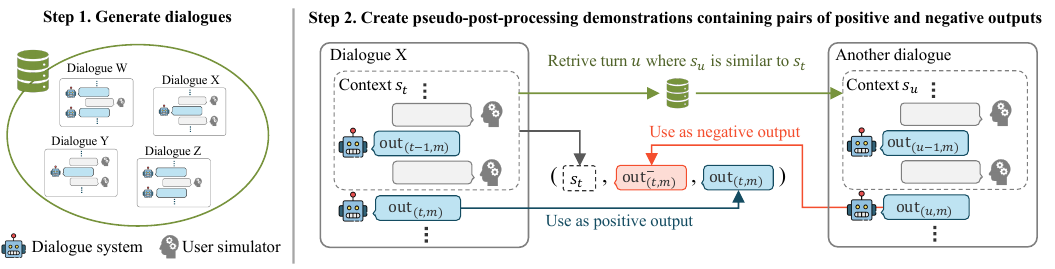}
\caption{Procedure for creating pseudo-post-processing demonstration data. First, we generate dialogues between the dialogue system and the user simulator. Subsequently, we create pairs of positive and negative outputs, where the output $\text{out}_t$ of module $m$ for context $s_t$ at turn $t$ is positive and the output $\text{out}_u$ at another turn $u$ is negative (i.e., $\text{out}_t^-$). In imitation learning stage, the reconstruction from $\text{out}_t^-$ to $\text{out}_t$ is learned as pseudo-post-processing.}
\label{fig:pseudo_pp_demo_creation}
\end{figure*}

\section{Preliminary}
\label{sec:preliminary}
The problem of learning capabilities for multi-turn task-oriented dialogue is often formulated as an MDP and optimized through RL. An MDP is defined by tuple $(\mathcal{S}, \mathcal{A}, \mathcal{P}, R, \gamma)$. Essentially, $\mathcal{S}$ and $\mathcal{A}$ represent all possible dialogue histories and system response sentences, respectively. $\mathcal{P}(s'|s,a)$ represents the transition model, $\mathcal{S} \times \mathcal{A} \times \mathcal{S} \rightarrow [0,1]$ defines the dialogue environment containing the user, and $R(s, a)$ represents the immediate reward function $\mathcal{S} \times \mathcal{A} \rightarrow \mathbb{R}$. $\gamma$ denotes the discount factor. At each turn $t$, the policy $\mathcal{F}: \mathcal{S} \rightarrow \mathcal{A}$ (i.e., the dialogue system) samples an action (i.e., the system response) $a_t \sim \mathcal{F}(a_t | s_t)$. Until the final state at turn $T$ is reached, the next state $s'_t \sim P(s'_t | s_t, a_t)$ and the immediate reward $r_t = R(s_t, a_t)$ are obtained. The goal of RL is to train $\mathcal{F}$ to maximize the value function $V$, which is the expected cumulative discounted reward as follows.
\small
\begin{equation}
\label{equ:value_function}
V^\mathcal{F}(s) := \mathbb{E}\left[\sum_{t=0}^T \gamma^t r_t | s_0 = s\right]
\end{equation}
\normalsize
Numerous studies targeted only a part of $\mathcal{F}$, such as the policy module, rather than the entire $\mathcal{F}$.

In complex problems, such as task-oriented dialogues, directly obtaining a policy that maximizes Eq. ~(\ref{equ:value_function}) is challenging. One effective method is the policy-gradient-based approach~\citep{NIPS1999_464d828b}, which directly improves the policy network $\mathcal{F}_\theta$ parameterized by $\theta$. According to the policy gradient theorem, gradient $\nabla_\mathcal{F} J(\theta)$ is expressed as follows:
\small
\begin{equation}
\label{equ:policy_gradient}
\nabla_\mathcal{F} J(\theta) = \mathbb{E} \left[\sum_{t=0}^{T} \Psi_t \nabla_\theta \log \mathcal{F}_\theta(a_t | s_t) \right]
\end{equation}
\normalsize
The specific definition of $\Psi_t$ varies depending on the implementation of the RL algorithm, such as the sum of the rewards obtained over all turns or advantage estimates~\citep{schulman2015high}. 

\section{Proposed Method}
In this section, we explain the problem formulation of our study, the proposed UniPPN, imitation learning (IL) and  RL, which together constitute our optimization procedure for UniPPN.

\subsection{Problem Formulation}
Here, we formulate the optimization problem for the dialogue system $\mathcal{F}$ through post-processing. We assume that $\mathcal{F}$ has a modularity consisting of $M$ modules: $\text{Module}_1$, ..., $\text{Module}_M$. At each turn $t$, each module $\text{Module}_m$ takes the output $\text{out}_{(t,m-1)}$ of the previous $\text{Module}_{m-1}$ as its input $\text{in}_{(t,m)}$, outputting its processing result $\text{out}_{(t,m)} \sim \text{Module}_m(\text{in}_{(t,m)})$. Some modules may use the dialogue history $s_t$ as additional input. The post-processing network $\text{PPN}_m$ for $\text{Module}_m$ modifies $\text{out}_{(t,m)}$ and the modified $\text{out}_{(t,m)}^+ \sim \text{PPN}_m(s_t, \text{in}_{(t,m)}, \text{out}_{(t,m)})$ becomes the input for $\text{Module}_{m+1}$. For the optimization of $\mathcal{F}$, we train $\text{PPN}_m$ instead of $\text{Module}_m$.

\subsection{UniPPN}
In our proposed method, the post-processing of the outputs from all $M$ modules is performed by a single network, UniPPN $\pi$ (Figure~\ref{fig:unippn}). Specifically, UniPPN modifies the output of any $\text{Module}_m$: $\text{out}_{(t,m)}^+$ $\sim$ $\text{UniPPN}(s_t,$ $\text{in}_{(t,m)},$ $\text{out}_{(t,m)},$ $\text{prefix}_m)$. Here, $\text{prefix}_m$ is an indicator that specifies that the module to be post-processed is $\text{Module}_m$. The input and output formats of UniPPN are text sequences, with post-processing executed as a sequence-transformation task. For the tokenized sequences $\bm{x}$ $=$ $(x_1$, ..., $x_k)$ and $\bm{y}$ $=$ $(y_1$, ..., $y_l)$, representing $(s_t, \text{in}_{(t,m)}, \text{out}_{(t,m)}, \text{prefix}_m)$ and $\text{out}_{(t,m)}^+$ respectively, the following conditional probability is modeled:
\small
\begin{equation}
\pi_\theta(\bm{y} | \bm{x}) = \prod_{i=1}^{l} \pi_\theta(y_i | \bm{x}, \bm{y}_{<i})
\end{equation}
\normalsize
where $\pi_\theta$ represents a pre-trained language model  parameterized by $\theta$. By treating not only the post-processing of natural language, such as the output of NLG modules but also structural data, such as the output of NLU or DST as a sequence-transformation task~\citep{JMLR:v21:20-074, Liang_Tian_Chen_Yu_2020}, UniPPN can uniformly perform post-processing across all modules. 

\subsection{Imitation Learning of Post-Processing}
Pre-trained language models are typically trained on web text and may not sufficiently possess the ability to modify the outputs of modules in task-oriented dialogue systems. Therefore, we conduct additional pre-training to teach the model $\pi_\theta$ the formats of input $(s_t, \text{in}_{(t,m)}, \text{out}_{(t,m)}, \text{prefix}_m)$ and output $\text{out}_{(t,m)}^+$ through supervised fine-tuning. In the general RL paradigm, IL conducted before online RL uses demonstration data, which consist of the action history of experts, such as humans. However, in our problem setting, demonstration data for post-processing the outputs of each module in the dialogue system $\mathcal{F}$ do not exist. Therefore, we automatically generate post-processing demonstration data and use them for supervised fine-tuning.

Using the procedure shown in Figure~\ref{fig:pseudo_pp_demo_creation}, we create pseudo-post-processing demonstration data for each $\text{Module}_m$. This process involves sampling dialogues by repeating interactions between $\mathcal{F}$ and the environment $\mathcal{P}$ to generate the input-output history $h_{(t,m)}$ $=$ $(s_t, \text{in}_{(t,m)}, \text{out}_{(t,m)})$ for each $\text{Module}_m$ at each turn $t$, resulting in $H_m$ $=$ $\{h_{(1,m)},$ $...,$ $h_{(|H_m|,m)}\}$. We now demonstrate the modification of $\text{out}_{(t,m)}$. Here, the label $\text{out}_{(t,m)}^+$, which represents the correct modification of $\text{out}_{(t,m)}$, cannot be created automatically. Under the assumption that the output $\text{out}_{(t,m)}$ of $\text{Module}_m$ is reasonably valid, we consider $\text{out}_{(t,m)}$ to be the target output after post-processing; we use $\text{out}_{(t,m)}^-$, randomly sampled from another turn $u$ (which may be from the same or a different dialogue) as the negative output that should be post-processed. This creates one demonstration instance $d_{(t,m)}$ $=$ $\{(s_t$, $\text{in}_{(t,m)},$ $\text{out}_{(t,m)}^-,$ $\text{prefix}_m),$ $\text{out}_{(t,m)}\}$, representing the modification from $\text{out}_{(t,m)}^-$ to $\text{out}_{(t,m)}$. We applied this pseudo-data creation process to all samples in $H_m$, resulting in the final demonstration dataset $D_m$ $=$ $\{d_{(1,m)}, ..., d_{(|H_m|,m)}\}$. In the following section, we describe the two techniques used to create $D_m$.

\paragraph{Sampling Realistic $\text{out}^-$}
If we sample a turn $u$ that is completely irrelevant to the context of $t$, it could introduce noise, causing $\pi$ to potentially learn to ignore $\text{out}_{(t,m)}^-$ rather than to modify it appropriately. To ensure that the mistakes are reasonable, we sample turns with contexts similar to $t$. Specifically, from the entire history excluding $h_{(t,m)}$ (i.e., $H_m \setminus \{h_{(t,m)}\}$), we extract the top few turns with the highest cosine similarity to the vector representation of the context $s_t$ in $h_{(t,m)}$, and randomly sample $h_{(u,m)}$ from the extracted turns. We use a general-purpose embedding model, such as E5~\citep{wang2022text} to vectorize the context.

\paragraph{Learning to Copy} In post-processing, it is not always necessary to modify the outputs; outputs without issues should be ``copied'' without modification. To reflect this, during the IL phase, we input the original $\text{out}_{(t,m)}$ into $\pi$ to ensure that modifications are not always required. In these cases, the target output is only the special token ``\texttt{copy}''. Specifically, demonstrations of such cases are $d_{(t,m)}$ $=$ $\{(s_t,$ $\text{in}_{(t,m)},$ $\text{out}_{(t,m)},$ $\text{prefix}_m),$ $\texttt{copy}\}$. This approach allows the model to explicitly learn whether post-processing is necessary, while also reducing the generation costs when post-processing is unnecessary. Whether each instance becomes a copy instance is determined randomly using copy ratio $\alpha \in [0,1]$, which is a hyperparameter.

\vskip\baselineskip
We update $\theta$ based on the maximum likelihood objective using the final dataset $D_{1:M} = [D_1;$ $...;$ $D_M]$, which combines the pseudo-post-processing data for all $M$ modules. The optimized parameters in this IL step are denoted by $\phi$.

\subsection{Optimization with Reinforcement Learning}
In the RL phase, we install the UniPPN $\pi_\phi$ obtained from the IL step into the dialogue system $\mathcal{F}$. Subsequently, let $\mathcal{F}$ interact repeatedly with the environment $\mathcal{P}$ over multiple turns and update $\phi$ based on these experiences using a policy-gradient-based approach. In typical task-oriented dialogue systems using online RL, only a single policy network (e.g., a policy module) operates per turn, and it is updated according to Eq. ~(\ref{equ:policy_gradient}). In contrast, our study involves a policy $\pi$ that acts $M$ times per turn, and outputs the system response as action $a$. Although each of the $M$ actions should have different gradients based on their individual advantages, Eq. ~(\ref{equ:policy_gradient}) treats them as having the same contributions. This can result in coarse rewards and learning instability.

Therefore, we extend the standard MDP described in Section~\ref{sec:preliminary} and introduce a module-level MDP, where the unit of time step is the ``post-processing of one module by $\pi$'' rather than the ``one turn response by $\mathcal{F}$''. Specifically, the value function to be maximized and policy gradient of $\pi_\phi$ are as follows:
\small
\begin{equation}
\label{equ:module_level_value_function}
V^\mathcal{\pi}(\bm{x}) := \mathbb{E}\left[\sum_{t=0}^T \sum_{m=1}^{M} \gamma^{(t+1)(m-1)} r_{(t,m)} | \bm{x}_{(0,1)} = \bm{x}\right]
\end{equation}
\begin{equation}
\label{equ:module_level_policy_gradient}
\nabla_\pi J(\phi) = \mathbb{E} \left[\sum_{t=0}^{T} \sum_{m=1}^M\Psi_{(t,m)} \nabla_\phi \log \pi_\phi(\bm{y}_{(t,m)} | \bm{x}_{(t,m)}) \right]
\end{equation}
\normalsize
Here, $r_{(t,m)}$ represents the immediate reward for post-processing the output of $\text{Module}_m$ at turn $t$. As in previous studies using online RL~\citep{hou-etal-2021-imperfect}, a small negative fixed value is assigned continuously until the end of the dialogue. $\bm{x}_{(t,m)}$ and $\bm{y}_{(t,m)}$ are the tokenized sequences of the input text $(s_t,$ $\text{in}_{(t,m)},$ $\text{out}_{(t,m)},$ $\text{prefix}_m)$ and output text of UniPPN, respectively. Eq.~(\ref{equ:module_level_policy_gradient}) shows that the gradients can be computed in $M$ gradient accumulation steps. Note that, in Eq.~(\ref{equ:module_level_value_function}), the number of calculations for the value function in the module-level MDP is $T \times M$, resulting in a possible exponential increase in the computational cost. However, because $M$ in a typical task-oriented dialogue system is four at most, this is not a problem in practice.

To implement $\Psi_{(t,m)}$, we adopt a generalized advantage estimation~\citep{schulman2015high}. Specifically, we compute the advantage estimate $\hat{A}_{(t,m)}$ based on the value $V_\psi (\bm{x}_{(t,m)})$ of $\bm{x}_{(t,m)}$ estimated using another language model $V_\psi$ parameterized by $\psi$ as a critic network:
\small
\begin{equation}
\label{equ:advantage_estimate}
\begin{split}
\hat{A}_{(t,m)} &= \delta_{(t,m)} + \gamma \lambda \hat{A}_{(t,m)'}, \\
\delta_{(t,m)} &= r_{(t,m)} + \gamma V_\psi(\bm{x}_{(t,m)'}) - V_\psi(\bm{x}_{(t,m)})
\end{split}
\end{equation}
\begin{equation}
\label{equ:module_level_timestep_increment}
(t,m)' =
\begin{cases}
(t,m+1) & \text{if } m < M \\
(t+1, 1) & \text{if } m = M
\end{cases}
\end{equation}
\normalsize
Here, $\delta_{(t,m)}$ represents the TD residual, and the hyperparameter $\lambda \in [0, 1]$ controls the trade-off between utilizing actual long-term rewards and the estimated values. Because $V_\psi$ estimates the state value for each $\text{Module}_m$ at each turn $t$, fine-grained advantage estimation according to the contribution of each module is possible even in settings with sparse rewards across multi-turn dialogues. An advantage of this algorithm is that it does not require a high-cost manual reward design for each module, as required in previous studies. $V_\psi$ is trained to minimize the mean squared error with respect to the cumulative reward and $\pi_\phi$ is optimized using a clipped surrogate objective with proximal policy optimization (PPO)~\citep{schulman2017proximal}. For a detailed implementation of the RL algorithm, refer to Appendix~\apdxref{appendix:sec:rl_algorithm}{A}.

\section{Experiments}
In this evaluation experiment, we demonstrate that joint optimization using UniPPN is more effective than disjoint optimization combining conventional BinPPN and GenPPN for post-processing outputs from all modules to improve task-oriented dialogue systems.

\subsection{Experimental Setup}
We conducted evaluation experiments using the MultiWOZ~\citep{budzianowski-etal-2018-multiwoz} dataset, which contains a multi-domain task-oriented dialogue on travel information between customers and clerks. We applied UniPPN to various dialogue systems developed for MultiWOZ and assessed their task completion capabilities. For the user simulation, we used the agenda-based user simulator~\citep{schatzmann-etal-2007-agenda} provided by ConvLab-2~\citep{zhu-etal-2020-convlab}, which is an evaluation toolkit for task-oriented dialogue systems.

The dialogue systems used in our experiments included both pipeline and end-to-end systems. For the modules constituting the pipeline systems, we selected relatively recently proposed models that ranked high on the MultiWOZ benchmark\footnote{\url{https://github.com/budzianowski/multiwoz}} and  had publicly available implementations. The models adopted for each module in the pipeline system are as follows:
\begin{description}
\item[NLU] BERT NLU~\citep{chen2019bert}, a classification model based on BERT~\citep{devlin-etal-2019-bert}.
\item[DST] Rule-based DST and D3ST~\citep{zhao2022description}, a state-of-the-art word-level DST based on T5~\citep{JMLR:v21:20-074}.
\item[Policy] Rule-based policy and PPO policy fine-tuned using PPO~\citep{schulman2017proximal}. We also used LAVA~\citep{lubis-etal-2020-lava}, a word-level policy.
\item[NLG] Template-based NLG and SC-GPT~\citep{peng-etal-2020-shot}, which is based on GPT-2~\citep{radford2019language}.
\end{description}
For end-to-end systems, we adopted two representative models: PPTOD~\citep{su-etal-2022-multi} and an LLM-based model~\citep{hudecek-dusek-2023-large}. PPTOD is a T5-based dialogue model that is fine-tuned using MultiWOZ. The LLM-based model performs word-level DST and a word-level policy based on in-context learning with few-shot examples retrieved from MultiWOZ. For the LLM, we used GPT-4o mini provided by OpenAI's API.\footnote{\url{https://platform.openai.com}}

\subsection{Evaluation Metrics}
In the evaluation, each dialogue system interacted with the user simulator 1,024 times, and each of the 1,024 different user goals for testing was set in each dialogue. We reported the average score of 1,024 dialogues as the final score.

As evaluation metrics, we used the average \emph{ turns } across all dialogues, which represents the number of turns required to achieve a task, with lower values indicating better efficiency. Each turn comprised a pair of a user utterance and the corresponding system response. We also used \emph{Inform Recall/Precision/F1}. These metrics assess whether the system responds adequately to the information requested by the user during a dialogue. In addition, we used the \emph{Goal Match Rate} to assess whether the conditions of the entity (specific facilities, such as hotels) presented by the system matched the user's goal. Similarly, we also assessed the conditions of the entity booked by the system using the \emph{Book Rate}. We set the maximum number of turns in one dialogue to 20. A task was only considered \emph{Success} if the Inform Recall, Match Rate, and Book Rate all reached 1.0 within 20 turns.

\subsection{UniPPN Implementation}
We used a medium-sized GPT-2~\citep{radford2019language} with 355M parameters as the backbone model for UniPPN. We chose this parameter size because of its superior balance between the computational cost and performance, which was confirmed through preliminary experiments. 

\paragraph{Imitation Learning}
To construct the post-processing demonstration data $D_{1:M}$ for each dialogue system, we sampled 10,000 turns of interaction between the dialogue system and user simulator.\footnote{10,000 turns equal approximately 1,000 dialogues, although the number of turns per dialogue depends on various factors, such as the system's capabilities.} To sample turns with similar contexts in the construction of $D_{1:M}$, we adopted GTE-base~\citep{li2023towards} as the embedding model and used the latest three utterances as the context. In addition, we set the copy ratio $\alpha$ to 0.1 throughout the experiment because, in preliminary experiments, we examined 9 levels from 0.1 to 0.9 for $\alpha$ and found that 0.1 yielded the highest reward.

\paragraph{Reinforcement Learning}
As an approximator for the value function, we used GPT-2 of 124M parameters, with an additional feedforward network outputting a scalar value. For the reward function $R(t,m)$, we set a small negative value of $R(t,m) = -0.1$ until the end of the dialogue and $R(T,M) =2 \times M$ for the final step if the task was achieved. We trained for 200 iterations, and in each iteration, we sampled 1,024 turns as training data. We used the checkpoints from the final iteration for testing.

\vskip\baselineskip
For detailed implementations and hyperparameters of the learning process, please refer to Appendix~\apdxref{appendix:sec:details_unippn_training}{B}.

\begin{table*}[t]
\centering
\footnotesize
{\tabcolsep=1.7mm
\begin{tabular}{lllllccccccc}
\toprule
\multirow{2}{*}{\textbf{System}} & \multicolumn{4}{c}{\textbf{Module combination}} & \textbf{Success} & \multicolumn{3}{c}{\textbf{Inform}} & \textbf{Book} & \textbf{Match} & \multirow{2}{*}{\textbf{Turns $\downarrow$}} \\
\cmidrule{2-5}\cmidrule{7-9}
 & \textbf{NLU} & \textbf{DST} & \textbf{Policy} & \textbf{NLG} & \textbf{Rate} & \textbf{Recall} & \textbf{Prec.} & \textbf{F1} & \textbf{Rate} & \textbf{Rate} &  \\
\midrule
\multicolumn{12}{l}{\emph{Pipeline System}} \\
\midrule
SYS$_\text{RULE}$ & BERT & Rule & Rule & Template & 83.69 & 93.72 & 81.38 & 85.08 & 91.19 & 91.63 & 5.92 \\
+BinPPN\&GenPPN & \checkmark & \checkmark & \checkmark & \checkmark & 85.25 & 93.96 & \textbf{83.53} & 86.54 & 91.18 & 92.68 & \textbf{5.67} \\
+UniPPN & \checkmark & \checkmark & \checkmark & \checkmark & \textbf{90.62} & \textbf{96.46} & 82.86 & \textbf{87.19} & \textbf{97.13} & \textbf{94.81} & 5.77 \\
\midrule
SYS$_\text{D3ST}$ & -- & D3ST$^{\text{word}}$ & Rule & Template & 58.50 & 72.41 & 67.36 & 66.62 & 62.22 & 77.12 & 6.78 \\
+BinPPN\&GenPPN & -- & \checkmark & \checkmark & \checkmark & 67.58 & 82.24 & 68.02 & 71.89 & 75.31 & 81.59 & 6.36 \\
+UniPPN & -- & \checkmark & \checkmark & \checkmark & \textbf{85.06} & \textbf{93.63} & \textbf{71.82} & \textbf{78.56} & \textbf{89.25} & \textbf{92.15} & \textbf{5.78} \\
\midrule
SYS$_\text{PPO}$ & BERT & Rule & PPO & Template & 69.24 & 86.90 & 66.99 & 72.90 & 80.74 & 83.79 & 8.55 \\
+BinPPN\&GenPPN & \checkmark & \checkmark & \checkmark & \checkmark & 70.61 & 86.59 & 66.08 & 72.46 & 80.71 & 84.42 & 8.46 \\
+UniPPN & \checkmark & \checkmark & \checkmark & \checkmark & \textbf{89.36} & \textbf{96.72} & \textbf{68.37} & \textbf{77.68} & \textbf{94.73} & \textbf{94.82} & \textbf{5.24} \\
\midrule
SYS$_\text{SCGPT}$ & BERT & Rule & Rule & SC-GPT & 75.29 & 93.73 & 79.49 & 83.92 & 68.53 & 92.06 & 6.12 \\
+BinPPN\&GenPPN & \checkmark & \checkmark & \checkmark & \checkmark & 86.72 & 94.66 & 82.79 & 86.32 & 92.53 & 92.84 & \textbf{5.68} \\
+UniPPN & \checkmark & \checkmark & \checkmark & \checkmark & \textbf{90.14} & \textbf{97.19} & \textbf{84.14} & \textbf{88.38} & \textbf{95.89} & \textbf{95.69} & 6.10 \\
\midrule
SYS$_\text{LAVA}$ & BERT & Rule & LAVA$^{\text{word}}$ & -- & 64.36 & 82.87 & 55.74 & 64.04 & 72.96 & 80.34 & 10.24 \\
+UniPPN & \checkmark & \checkmark & \checkmark & -- & \textbf{79.39} & \textbf{98.10} & \textbf{64.41} & \textbf{75.21} & \textbf{88.66} & \textbf{89.91} & \textbf{5.84} \\
\midrule
\multicolumn{12}{l}{\emph{End-to-End System}} \\
\midrule
SYS$_\text{PPTOD}$ & -- & PPTOD$^{\text{word}}$ & PPTOD$^{\text{word}}$ & -- & 61.04 & 82.19 & 75.12 & 76.07 & 50.80 & 83.61 & 8.15 \\
+UniPPN & -- & \checkmark & \checkmark & -- & \textbf{80.37} & \textbf{92.71} & \textbf{75.83} & \textbf{81.00} & \textbf{83.08} & \textbf{90.01} & \textbf{6.97} \\
\midrule
SYS$_\text{LLM}$ (GPT-4o mini) & -- & LLM$^{\text{word}}$ & LLM$^{\text{word}}$ & -- & 62.30 & 77.66 & 58.64 & 62.85 & 62.37 & 79.26 & 7.66 \\
+UniPPN & -- & \checkmark & \checkmark & -- & \textbf{88.28} & \textbf{94.80} & \textbf{65.01} & \textbf{74.31} & \textbf{90.81} & \textbf{93.57} & \textbf{4.66} \\
\bottomrule
\end{tabular}
}
\caption{Test results for each dialogue system and when post-processing is applied to all of their modules using BinPPN and GenPPN (+BinPPN\&GenPPN) or using UniPPN (+UniPPN). The \checkmark under each module in the row indicates whether PPN is applied to that module. Note that UniPPNs applied to the same system are the same network. The superscript ``$\text{word}$'' for DST and policy indicates that they are word-level DST and word-level policy, respectively.}
\label{tab:result_ppn_all}
\vspace{-2mm}
\end{table*}

\subsection{Baselines}
As baselines for this experiment, we used two methods: the original dialogue system without post-processing and a method that post-processes the outputs of all modules by combining conventional BinPPN and GenPPN (called BinPPN\&GenPPN). Because BinPPN and GenPPN cannot be trained jointly, RL was used to train the two types of PPNs (BinPPN and GenPPN) separately. Specifically, we used RL to optimize the post-processing of the three modules (NLU, DST, and policy) using BinPPN. Thereafter, while installing these three BinPPNs in the system and freezing their parameters, we attached GenPPN to NLG and trained it again using RL. The BinPPN was optimized first because NLU, DST, and policy, whose output BinPPN post-processes, precede NLG, whose output GenPPN post-processes; therefore, this order of optimization is considered appropriate. Note that, contrary to UniPPN, BinPPN\&GenPPN require two RL phases. For the implementation and hyperparameters of BinPPN and GenPPN, we used those published and reported in previous studies~\citep{ohashi-higashinaka-2022-post, ohashi-higashinaka-2023-enhancing}. However, for the backbone model of GenPPN,  we used Llama 3.1 8B~\citep{dubey2024llama} instead of Llama 7B~\citep{touvron2023llama} as described in the previous study. 

\subsection{Main Results}
Table~\ref{tab:result_ppn_all} shows the test results when post-processing was applied to all modules of each system. We evaluated two cases for the application of post-processing to pipeline systems consisting of multiple modules: +BinPPN\&GenPPN and +UniPPN. Because BinPPN\&GenPPN cannot be applied to SYS$_{\text{LAVA}}$ or end-to-end systems that do not output DAs, only UniPPN was applied to these systems.

In the pipeline systems, +UniPPN significantly outperformed +BinPPN\&GenPPN. For example, even for systems, such as SYS$_{\text{RULE}}$ and SYS$_{\text{PPO}}$, where performance improvement by +BinPPN\&GenPPN was limited, UniPPN improved the performance. In particular, it is noteworthy that UniPPN enhanced SYS$_{\text{RULE}}$ to 90.62 points, considering that SYS$_{\text{RULE}}$ comprises high-performance modules that are carefully crafted by hand-written rules, and its original success rate is high at approximately 84 points. Furthermore, we applied UniPPN to the system, including the word-level policy and end-to-end systems, to which conventional BinPPN or GenPPN could not be applied. UniPPN significantly improved the original performance of SYS$_{\text{LAVA}}$, SYS$_{\text{PPTOD}}$, and SYS$_{\text{LLM}}$ for all evaluation metrics.

From these results, we demonstrated that the joint optimization of post-processing the outputs from all modules using UniPPN is more effective than the disjoint optimization of multiple PPNs. Furthermore, its efficacy was demonstrated, and it was shown that it can be applied to any dialogue system, regardless of the trainability of each module, including rule- or API-based modules. We believe that UniPPN is more practical considering the high training cost of BinPPN\&GenPPN and the complexity of installing multiple networks in the systems.

\subsection{Comparison with either BinPPN or GenPPN}

\begin{table}[t]
\centering
\footnotesize
{\tabcolsep=1.3mm
\begin{tabular}{llccccc}
\toprule
\textbf{System} & \textbf{PPN} & \textbf{Success} & \textbf{Inf. F1} & \textbf{Book} & \textbf{Match} & \textbf{Turns $\downarrow$} \\
\midrule
SYS$_\text{RULE}$ & Bin & 83.40 & \textbf{86.07} & 91.13 & 91.47 & 6.04 \\
 & Uni & \textbf{87.11} & 83.81 & \textbf{96.08} & \textbf{93.13} & \textbf{5.80} \\
\midrule
SYS$_\text{D3ST}$ & Bin & 62.89 & 70.78 & 70.71 & 79.02 & 7.21 \\
 & Uni & \textbf{77.93} & \textbf{76.97} & \textbf{79.50} & \textbf{88.66} & \textbf{5.83} \\
\midrule
SYS$_\text{PPO}$ & Bin & 69.82 & 72.80 & 80.78 & 83.64 & 8.63 \\
 & Uni & \textbf{85.84} & \textbf{76.37} & \textbf{94.05} & \textbf{92.50} & \textbf{5.40} \\
\midrule
SYS$_\text{SCGPT}$ & Bin & 73.44 & 83.60 & 63.06 & 91.81 & \textbf{6.08} \\
 & Uni & \textbf{77.44} & \textbf{85.06} & \textbf{69.34} & \textbf{93.62} & 6.10 \\
\midrule
SYS$_\text{LAVA}$ & Bin & 64.06 & \textbf{69.56} & \textbf{80.99} & 79.98 & \textbf{8.72} \\
 & Uni & \textbf{65.72} & 67.55 & 76.48 & \textbf{80.83} & 9.45 \\
\bottomrule
\end{tabular}
}
\caption{Test results when applying either BinPPN or UniPPN to the three modules of NLU, DST, and policy in each system. For SYS$_\text{LAVA}$, which uses word-level policy, BinPPN cannot be applied, therefore, either BinPPN or UniPPN is applied only to NLU and DST.}
\label{tab:result_ppn_utp}
\end{table}

\begin{table}[t]
\centering
\footnotesize
{\tabcolsep=1.3mm
\begin{tabular}{llccccc}
\toprule
\textbf{System} & \textbf{PPN} & \textbf{Success} & \textbf{Inf. F1} & \textbf{Book} & \textbf{Match} & \textbf{Turns $\downarrow$} \\
\midrule
SYS$_\text{RULE}$ & Gen & 85.06 & 85.08 & 90.90 & \textbf{92.35} & \textbf{5.61} \\
 & Uni & 85.06 & 85.08 & \textbf{91.58} & 92.29 & 5.81 \\
\midrule
SYS$_\text{SCGPT}$ & Gen & 84.08 & \textbf{85.40} & \textbf{91.62} & 91.65 & \textbf{5.67} \\
 & Uni & \textbf{84.77} & 81.66 & 90.50 & \textbf{92.56} & 6.15 \\
\bottomrule
\end{tabular}
}
\caption{Test results when applying either GenPPN or UniPPN only to the NLG of each system.}
\label{tab:result_ppn_g}
\vspace{-2mm}
\end{table}

To clarify the fundamental performance difference between UniPPN and BinPPN, we compared the performance when either BinPPN or UniPPN was applied to NLU, DST, and policy, which BinPPN can be applied, for each system. Table~\ref{tab:result_ppn_utp} presents the results. We can observe that for most systems, the performance improvement by UniPPN exceeds that of BinPPN. This performance difference may be due to limited post-processing capabilities of BinPPN, which is limited to basic binary operations, leaving little room for improvement, whereas UniPPN can flexibly generate various types of information.
Similarly, we compared the performance when either GenPPN or UniPPN was applied to the NLG of each system. Table~\ref{tab:result_ppn_g} shows the results. For the turn metric, GenPPN consistently outperformed UniPPN. This could be because GenPPN learns to generate responses, such that DAs are easily understood by the user simulator, thereby reducing the increase in turns caused by users asking back. However, for other metrics, such as success rate, there was no significant difference between the two methods. Considering that the training algorithm of GenPPN requires the internal DAs of the system and feedback on whether the user understood those DAs, UniPPN, which only requires the final dialogue outcome as a reward, is promising.

Based on  these results, we demonstrated that UniPPN addresses the multiple challenges of conventional BinPPN and GenPPN, resulting in improved task completion capability, reduced computational cost, and broader applicability.  

\section{Human Evaluation}

\begin{table}[t]
\centering
\footnotesize
{\tabcolsep=1.3mm
\begin{tabular}{lcclccc}
\toprule
\textbf{System} & $N$ & \textbf{Success} & \textbf{Turns $\downarrow$} & \textbf{LU} & \textbf{RA} & \textbf{OS} \\
\midrule
SYS$_\text{PPO}$ & 43 & 46.51 & 8.79 & 2.70 & 2.70	 & \textbf{2.63} \\
+BinPPN\&GenPPN & 40 & 42.50 & 9.80 & 2.70 & 2.85 & 2.38 \\
+UniPPN & 48 & \textbf{54.17} & \textbf{8.06$^+$} & \textbf{2.81} & \textbf{2.90} & 2.56 \\
\bottomrule
\end{tabular}
}
\caption{Results of human evaluation. $N$ indicates the number of subjects who conversed with each system. LU, RA, and OS indicate evaluations of the system's language understanding, response appropriateness, and overall satisfaction, respectively. $+$ indicates that there was a significant tendency with $p<0.1$ according to the Mann-Whitney U test in the difference between the scores of +BinPPN\&GenPPN and +UniPPN.}
\label{tab:result_human_eval}
\vspace{-2mm}
\end{table}

We verified whether the performance improvement of dialogue systems by UniPPN is also effective for human users. Specifically, for SYS$_\text{PPO}$ in Table~\ref{tab:result_ppn_all}, we had human users interact with three types of systems (i.e., the original system, +BinPPN\&GenPPN, and +UniPPN) and evaluated their performance. We chose SYS$_\text{PPO}$ because it achieved the lowest task success rate among the systems containing all four modules. We believe that the impact of post-processing on the performance is most apparent in this case. We recruited more than 40 workers for each system on Amazon Mechanical Turk (AMT) and had them interact with one of the three systems to achieve dialogue goals randomly created for each dialogue. After the dialogue, the workers were requested to subjectively evaluate the system's language understanding (LU), response appropriateness (RA), and overall satisfaction (OS) with the dialogue on a 5-point Likert scale. Ethical approval was obtained from our institution before the experiment. For detailed experimental settings, please refer to Appendix~\apdxref{appendix:sec:human_evaluation_detail}{C}.

Table~\ref{tab:result_human_eval} lists the results of the human evaluation metrics for the task completion and subjective assessments. For the evaluation metrics related to task completion capability, such as success rate and number of turns, +UniPPN outperformed both the original SYS$_\text{PPO}$ and +BinPPN\&GenPPN. Notably, the difference in the number of turns showed a statistically significant tendency, highlighting the effectiveness of UniPPN, which can jointly learn to post-process the outputs of all modules, compared with disjointly trained conventional PPNs. By contrast, +BinPPN\&GenPPN performed worse than the original SYS$_\text{PPO}$ for most metrics. This may be because the post-processing for each module is trained disjointly, preventing coordination between PPNs and deteriorating the overall system performance. Regarding the subjective evaluation metrics of LU, RA, and OS, there were no significant differences between SYS$_\text{PPO}$ and +UniPPN. This was expected, considering that UniPPN's reward signals were only related to task completion and did not include signals related to dialogue satisfaction.

\section{Conclusion}
In this study, we proposed UniPPN, a method that jointly learns the post-processing of outputs from all modules in task-oriented dialogue systems. Using simulation experiments based on the MultiWOZ dataset, we applied UniPPN to various pipeline systems with recent high-performance modules and end-to-end systems, including a GPT-4o mini-powered system. Our results confirm that UniPPN significantly outperforms conventional PPN-based methods in terms of task completion performance of dialogue systems. Furthermore, human evaluation experiments demonstrated that UniPPN, optimized in a simulation environment, is effective in real-world scenarios.

In future studies, we aim to reduce the overall training cost of UniPPN. This involves reducing the number of dialogue experiences required for learning convergence and optimizing the model size for efficiency. In addition, we plan to extend beyond text dialogue to support the post-processing of outputs from modules in multimodal dialogue systems.



\section*{Acknowledgments}
This work was supported by JST Moonshot R\&D, Grant number JPMJMS2011. We used the computational resources of the supercomputer ``Flow'' at the Information Technology Center, Nagoya University.

\bibliography{references}

\begin{thebibliography}{40}
\providecommand{\natexlab}[1]{#1}

\bibitem[{Budzianowski et~al.(2018)Budzianowski, Wen, Tseng, Casanueva, Ultes, Ramadan, and Ga{\v{s}}i{\'c}}]{budzianowski-etal-2018-multiwoz}
Budzianowski, P.; Wen, T.-H.; Tseng, B.-H.; Casanueva, I.; Ultes, S.; Ramadan, O.; and Ga{\v{s}}i{\'c}, M. 2018.
\newblock {MultiWOZ - A Large-Scale Multi-Domain Wizard-of-Oz Dataset for Task-Oriented Dialogue Modelling}.
\newblock In \emph{Proceedings of the 2018 Conference on Empirical Methods in Natural Language Processing}, 5016--5026.

\bibitem[{Chen, Zhuo, and Wang(2019)}]{chen2019bert}
Chen, Q.; Zhuo, Z.; and Wang, W. 2019.
\newblock {BERT for Joint Intent Classification and Slot Filling}.
\newblock \emph{arXiv preprint arXiv:1902.10909}.

\bibitem[{Chen et~al.(2022)Chen, Chen, Zhou, and Yu}]{10.1007/978-981-99-2401-1_25}
Chen, Z.; Chen, L.; Zhou, X.; and Yu, K. 2022.
\newblock {Deep Reinforcement Learning for On-line Dialogue State Tracking}.
\newblock In \emph{Proceedings of the 17th National Conference on Man–Machine Speech Communication}, 278--292.

\bibitem[{Deng et~al.(2024)Deng, Zhang, Lam, Ng, and Chua}]{deng2024plugandplay}
Deng, Y.; Zhang, W.; Lam, W.; Ng, S.-K.; and Chua, T.-S. 2024.
\newblock Plug-and-Play Policy Planner for Large Language Model Powered Dialogue Agents.
\newblock In \emph{Proceedings of the Twelfth International Conference on Learning Representations}.

\bibitem[{Devlin et~al.(2019)Devlin, Chang, Lee, and Toutanova}]{devlin-etal-2019-bert}
Devlin, J.; Chang, M.-W.; Lee, K.; and Toutanova, K. 2019.
\newblock {BERT: Pre-training of Deep Bidirectional Transformers for Language Understanding}.
\newblock In \emph{Proceedings of the 2019 Conference of the North {A}merican Chapter of the Association for Computational Linguistics: Human Language Technologies}, 4171--4186.

\bibitem[{Dubey et~al.(2024)Dubey, Jauhri, Pandey, Kadian, Al-Dahle, Letman, Mathur, Schelten, Yang, Fan et~al.}]{dubey2024llama}
Dubey, A.; Jauhri, A.; Pandey, A.; Kadian, A.; Al-Dahle, A.; Letman, A.; Mathur, A.; Schelten, A.; Yang, A.; Fan, A.; et~al. 2024.
\newblock The Llama 3 Herd of Models.
\newblock \emph{arXiv preprint arXiv:2407.21783}.

\bibitem[{He et~al.(2022)He, Dai, Zheng, Wu, Cao, Liu, Jiang, Yang, Huang, Si, Sun, and Li}]{He_Dai_Zheng_Wu_Cao_Liu_Jiang_Yang_Huang_Si_Sun_Li_2022}
He, W.; Dai, Y.; Zheng, Y.; Wu, Y.; Cao, Z.; Liu, D.; Jiang, P.; Yang, M.; Huang, F.; Si, L.; Sun, J.; and Li, Y. 2022.
\newblock GALAXY: A Generative Pre-trained Model for Task-Oriented Dialog with Semi-supervised Learning and Explicit Policy Injection.
\newblock In \emph{Proceedings of 36th AAAI Conference on Artificial Intelligence}, 10, 10749--10757.

\bibitem[{Hosseini-Asl et~al.(2020)Hosseini-Asl, McCann, Wu, Yavuz, and Socher}]{NEURIPS2020_e9462095}
Hosseini-Asl, E.; McCann, B.; Wu, C.-S.; Yavuz, S.; and Socher, R. 2020.
\newblock A Simple Language Model for Task-Oriented Dialogue.
\newblock In \emph{Proceedings of Advances in Neural Information Processing Systems}, 20179--20191.

\bibitem[{Hou et~al.(2021)Hou, Liu, Zhao, Ou, Liu, Chen, and Zheng}]{hou-etal-2021-imperfect}
Hou, Z.; Liu, B.; Zhao, R.; Ou, Z.; Liu, Y.; Chen, X.; and Zheng, Y. 2021.
\newblock {Imperfect also Deserves Reward: Multi-Level and Sequential Reward Modeling for Better Dialog Management}.
\newblock In \emph{Proceedings of the 2021 Conference of the North American Chapter of the Association for Computational Linguistics: Human Language Technologies}, 2993--3001.

\bibitem[{Hude{\v{c}}ek and Dusek(2023)}]{hudecek-dusek-2023-large}
Hude{\v{c}}ek, V.; and Dusek, O. 2023.
\newblock {Are Large Language Models All You Need for Task-Oriented Dialogue?}
\newblock In \emph{Proceedings of the 24th Meeting of the Special Interest Group on Discourse and Dialogue}, 216--228.

\bibitem[{Kwan et~al.(2023)Kwan, Wang, Wang, and Wong}]{kwan2023survey}
Kwan, W.-C.; Wang, H.-R.; Wang, H.-M.; and Wong, K.-F. 2023.
\newblock A survey on recent advances and challenges in reinforcement learning methods for task-oriented dialogue policy learning.
\newblock \emph{Machine Intelligence Research}, 20(3): 318--334.

\bibitem[{Li et~al.(2020)Li, Lee, Peng, Li, Kiseleva, de~Rijke, Shayandeh, and Gao}]{li-etal-2020-guided}
Li, Z.; Lee, S.; Peng, B.; Li, J.; Kiseleva, J.; de~Rijke, M.; Shayandeh, S.; and Gao, J. 2020.
\newblock {Guided Dialogue Policy Learning without Adversarial Learning in the Loop}.
\newblock In \emph{Findings of the 2019 Conference on Empirical Methods in Natural Language Processing}, 2308--2317.

\bibitem[{Li et~al.(2023)Li, Zhang, Zhang, Long, Xie, and Zhang}]{li2023towards}
Li, Z.; Zhang, X.; Zhang, Y.; Long, D.; Xie, P.; and Zhang, M. 2023.
\newblock Towards general text embeddings with multi-stage contrastive learning.
\newblock \emph{arXiv preprint arXiv:2308.03281}.

\bibitem[{Liang et~al.(2020)Liang, Tian, Chen, and Yu}]{Liang_Tian_Chen_Yu_2020}
Liang, W.; Tian, Y.; Chen, C.; and Yu, Z. 2020.
\newblock {MOSS: End-to-End Dialog System Framework with Modular Supervision}.
\newblock In \emph{Proceedings of the 34th AAAI Conference on Artificial Intelligence}, 8327--8335.

\bibitem[{Liu et~al.(2018)Liu, T{\"u}r, Hakkani-T{\"u}r, Shah, and Heck}]{liu-etal-2018-dialogue}
Liu, B.; T{\"u}r, G.; Hakkani-T{\"u}r, D.; Shah, P.; and Heck, L. 2018.
\newblock {Dialogue Learning with Human Teaching and Feedback in End-to-End Trainable Task-Oriented Dialogue Systems}.
\newblock In \emph{Proceedings of the 2018 Conference of the North {A}merican Chapter of the Association for Computational Linguistics: Human Language Technologies}, 2060--2069.

\bibitem[{Lubis et~al.(2020)Lubis, Geishauser, Heck, Lin, Moresi, van Niekerk, and Gasic}]{lubis-etal-2020-lava}
Lubis, N.; Geishauser, C.; Heck, M.; Lin, H.-c.; Moresi, M.; van Niekerk, C.; and Gasic, M. 2020.
\newblock {LAVA: Latent Action Spaces via Variational Auto-encoding for Dialogue Policy Optimization}.
\newblock In \emph{Proceedings of the 28th International Conference on Computational Linguistics}, 465--479.

\bibitem[{Ni et~al.(2022)Ni, Young, Pandelea, Xue, and Cambria}]{ni2022recent}
Ni, J.; Young, T.; Pandelea, V.; Xue, F.; and Cambria, E. 2022.
\newblock Recent advances in deep learning based dialogue systems: A systematic survey.
\newblock \emph{Artificial intelligence review}, 1--101.

\bibitem[{Ohashi and Higashinaka(2022)}]{ohashi-higashinaka-2022-post}
Ohashi, A.; and Higashinaka, R. 2022.
\newblock Post-processing Networks: Method for Optimizing Pipeline Task-oriented Dialogue Systems using Reinforcement Learning.
\newblock In \emph{Proceedings of the 23rd Annual Meeting of the Special Interest Group on Discourse and Dialogue}, 1--13.

\bibitem[{Ohashi and Higashinaka(2023)}]{ohashi-higashinaka-2023-enhancing}
Ohashi, A.; and Higashinaka, R. 2023.
\newblock {Enhancing Task-oriented Dialogue Systems with Generative Post-processing Network}.
\newblock In \emph{Proceedings of the 2023 Conference on Empirical Methods in Natural Language Processing}, 3815--3828.

\bibitem[{Peng et~al.(2020)Peng, Zhu, Li, Li, Li, Zeng, and Gao}]{peng-etal-2020-shot}
Peng, B.; Zhu, C.; Li, C.; Li, X.; Li, J.; Zeng, M.; and Gao, J. 2020.
\newblock {Few-shot Natural Language Generation for Task-Oriented Dialog}.
\newblock In \emph{Findings of the 2021 Conference on Empirical Methods in Natural Language Processing}, 172--182.

\bibitem[{Qin et~al.(2023)Qin, Pan, Chen, Liao, Yu, Zhang, Che, and Li}]{qin-etal-2023-end}
Qin, L.; Pan, W.; Chen, Q.; Liao, L.; Yu, Z.; Zhang, Y.; Che, W.; and Li, M. 2023.
\newblock {End-to-end Task-oriented Dialogue: A Survey of Tasks, Methods, and Future Directions}.
\newblock In \emph{Proceedings of the 2023 Conference on Empirical Methods in Natural Language Processing}, 5925--5941.

\bibitem[{Radford et~al.(2019)Radford, Wu, Child, Luan, Amodei, Sutskever et~al.}]{radford2019language}
Radford, A.; Wu, J.; Child, R.; Luan, D.; Amodei, D.; Sutskever, I.; et~al. 2019.
\newblock Language models are unsupervised multitask learners.
\newblock \emph{OpenAI blog}, 1(8): 9.

\bibitem[{Raffel et~al.(2020)Raffel, Shazeer, Roberts, Lee, Narang, Matena, Zhou, Li, and Liu}]{JMLR:v21:20-074}
Raffel, C.; Shazeer, N.; Roberts, A.; Lee, K.; Narang, S.; Matena, M.; Zhou, Y.; Li, W.; and Liu, P.~J. 2020.
\newblock Exploring the Limits of Transfer Learning with a Unified Text-to-Text Transformer.
\newblock \emph{Journal of Machine Learning Research}, 1--67.

\bibitem[{Schatzmann et~al.(2007)Schatzmann, Thomson, Weilhammer, Ye, and Young}]{schatzmann-etal-2007-agenda}
Schatzmann, J.; Thomson, B.; Weilhammer, K.; Ye, H.; and Young, S. 2007.
\newblock {Agenda-Based User Simulation for Bootstrapping a POMDP Dialogue System}.
\newblock In \emph{Proceedings of Human Language Technologies 2007: The Conference of the North {A}merican Chapter of the Association for Computational Linguistics}, 149--152.

\bibitem[{Schulman et~al.(2015)Schulman, Moritz, Levine, Jordan, and Abbeel}]{schulman2015high}
Schulman, J.; Moritz, P.; Levine, S.; Jordan, M.; and Abbeel, P. 2015.
\newblock {High-Dimensional Continuous Control Using Generalized Advantage Estimation}.
\newblock \emph{arXiv preprint arXiv:1506.02438}.

\bibitem[{Schulman et~al.(2017)Schulman, Wolski, Dhariwal, Radford, and Klimov}]{schulman2017proximal}
Schulman, J.; Wolski, F.; Dhariwal, P.; Radford, A.; and Klimov, O. 2017.
\newblock {Proximal Policy Optimization Algorithms}.
\newblock \emph{arXiv preprint arXiv:1707.06347}.

\bibitem[{Stiennon et~al.(2020)Stiennon, Ouyang, Wu, Ziegler, Lowe, Voss, Radford, Amodei, and Christiano}]{NEURIPS2020_1f89885d}
Stiennon, N.; Ouyang, L.; Wu, J.; Ziegler, D.; Lowe, R.; Voss, C.; Radford, A.; Amodei, D.; and Christiano, P.~F. 2020.
\newblock {Learning to summarize with human feedback}.
\newblock In \emph{Proceedings of Advances in Neural Information Processing Systems}, 3008--3021.

\bibitem[{Su et~al.(2022)Su, Shu, Mansimov, Gupta, Cai, Lai, and Zhang}]{su-etal-2022-multi}
Su, Y.; Shu, L.; Mansimov, E.; Gupta, A.; Cai, D.; Lai, Y.-A.; and Zhang, Y. 2022.
\newblock Multi-Task Pre-Training for Plug-and-Play Task-Oriented Dialogue System.
\newblock In \emph{Proceedings of the 60th Annual Meeting of the Association for Computational Linguistics}, 4661--4676.

\bibitem[{Sutton et~al.(1999)Sutton, McAllester, Singh, and Mansour}]{NIPS1999_464d828b}
Sutton, R.~S.; McAllester, D.; Singh, S.; and Mansour, Y. 1999.
\newblock {Policy Gradient Methods for Reinforcement Learning with Function Approximation}.
\newblock In \emph{Proceedings of Advances in Neural Information Processing Systems}, 1057--1063.

\bibitem[{Touvron et~al.(2023)Touvron, Lavril, Izacard, Martinet, Lachaux, Lacroix, Rozi{\`e}re, Goyal, Hambro, Azhar et~al.}]{touvron2023llama}
Touvron, H.; Lavril, T.; Izacard, G.; Martinet, X.; Lachaux, M.-A.; Lacroix, T.; Rozi{\`e}re, B.; Goyal, N.; Hambro, E.; Azhar, F.; et~al. 2023.
\newblock {LLaMA: Open and efficient foundation language models}.
\newblock \emph{arXiv preprint arXiv:2302.13971}.

\bibitem[{Tseng et~al.(2021)Tseng, Dai, Kreyssig, and Byrne}]{tseng-etal-2021-transferable}
Tseng, B.-H.; Dai, Y.; Kreyssig, F.; and Byrne, B. 2021.
\newblock {Transferable Dialogue Systems and User Simulators}.
\newblock In \emph{Proceedings of the 59th Annual Meeting of the Association for Computational Linguistics and the 11th International Joint Conference on Natural Language Processing}, 152--166.

\bibitem[{Wang et~al.(2020)Wang, Tian, Wang, Quan, and Yu}]{wang-etal-2020-multi-domain}
Wang, K.; Tian, J.; Wang, R.; Quan, X.; and Yu, J. 2020.
\newblock {Multi-Domain Dialogue Acts and Response Co-Generation}.
\newblock In \emph{Proceedings of the 58th Annual Meeting of the Association for Computational Linguistics}, 7125--7134.

\bibitem[{Wang et~al.(2022)Wang, Yang, Huang, Jiao, Yang, Jiang, Majumder, and Wei}]{wang2022text}
Wang, L.; Yang, N.; Huang, X.; Jiao, B.; Yang, L.; Jiang, D.; Majumder, R.; and Wei, F. 2022.
\newblock Text embeddings by weakly-supervised contrastive pre-training.
\newblock \emph{arXiv preprint arXiv:2212.03533}.

\bibitem[{Wu et~al.(2019)Wu, Madotto, Hosseini-Asl, Xiong, Socher, and Fung}]{wu-etal-2019-transferable}
Wu, C.-S.; Madotto, A.; Hosseini-Asl, E.; Xiong, C.; Socher, R.; and Fung, P. 2019.
\newblock {Transferable Multi-Domain State Generator for Task-Oriented Dialogue Systems}.
\newblock In \emph{Proceedings of the 57th Annual Meeting of the Association for Computational Linguistics}, 808--819.

\bibitem[{Wu et~al.(2023)Wu, Gung, Shu, and Zhang}]{wu-etal-2023-diacttod}
Wu, Q.; Gung, J.; Shu, R.; and Zhang, Y. 2023.
\newblock {D}iact{TOD}: Learning Generalizable Latent Dialogue Acts for Controllable Task-Oriented Dialogue Systems.
\newblock In \emph{Proceedings of the 24th Annual Meeting of the Special Interest Group on Discourse and Dialogue}, 255--267.

\bibitem[{Zhang et~al.(2020)Zhang, Takanobu, Zhu, Huang, and Zhu}]{zhang2020recent}
Zhang, Z.; Takanobu, R.; Zhu, Q.; Huang, M.; and Zhu, X. 2020.
\newblock {Recent advances and challenges in task-oriented dialog systems}.
\newblock \emph{Science China Technological Sciences}, 63(10): 1--17.

\bibitem[{Zhao et~al.(2022)Zhao, Gupta, Cao, Yu, Wang, Lee, Rastogi, Shafran, and Wu}]{zhao2022description}
Zhao, J.; Gupta, R.; Cao, Y.; Yu, D.; Wang, M.; Lee, H.; Rastogi, A.; Shafran, I.; and Wu, Y. 2022.
\newblock Description-driven task-oriented dialog modeling.
\newblock \emph{arXiv preprint arXiv:2201.08904}.

\bibitem[{Zhao and Eskenazi(2016)}]{zhao-eskenazi-2016-towards}
Zhao, T.; and Eskenazi, M. 2016.
\newblock {Towards End-to-End Learning for Dialog State Tracking and Management using Deep Reinforcement Learning}.
\newblock In \emph{Proceedings of the 17th Annual Meeting of the Special Interest Group on Discourse and Dialogue}, 1--10.

\bibitem[{Zhu et~al.(2020)Zhu, Zhang, Fang, Li, Takanobu, Li, Peng, Gao, Zhu, and Huang}]{zhu-etal-2020-convlab}
Zhu, Q.; Zhang, Z.; Fang, Y.; Li, X.; Takanobu, R.; Li, J.; Peng, B.; Gao, J.; Zhu, X.; and Huang, M. 2020.
\newblock {ConvLab-2: An Open-Source Toolkit for Building, Evaluating, and Diagnosing Dialogue Systems}.
\newblock In \emph{Proceedings of the 58th Annual Meeting of the Association for Computational Linguistics: System Demonstrations}, 142--149.

\bibitem[{Ziegler et~al.(2019)Ziegler, Stiennon, Wu, Brown, Radford, Amodei, Christiano, and Irving}]{ziegler2019fine}
Ziegler, D.~M.; Stiennon, N.; Wu, J.; Brown, T.~B.; Radford, A.; Amodei, D.; Christiano, P.; and Irving, G. 2019.
\newblock {Fine-tuning language models from human preferences}.
\newblock \emph{arXiv preprint arXiv:1909.08593}.

\end{thebibliography}

\ifshowapdx
    \clearpage
    \appendix

\part*{\centering Appendix}

\section{Reinforcement Learning Algorithm}
\label{appendix:sec:rl_algorithm}

In the implementation of our proposed method, we used Proximal Policy Optimization (PPO)~\citep{schulman2017proximal} as the optimization algorithm for $\pi$. In addition, we employed the generalized advantage estimate ~\citep{schulman2015high} to implement $\Psi_{(t,m)}$. The value network $V_\psi$ is trained to minimize the mean squared error with a cumulative reward:
\begin{equation}
\label{equ:value_loss}
\mathcal{L}_V(\psi) = \left( V_\psi(\bm{x}_{(t,m)}) - \sum_{i=0}^T\sum_{j=1}^M \gamma^{(i+1)(j-1)}r_{(t,m)+(i,j)} \right)^2
\end{equation}
Here, $(t,m)+(i,j)$ represents the progression of timesteps in the module-level MDP, similar to Eq. ~(\ref{equ:module_level_timestep_increment}). $V_\psi$ estimates the state value of each module $m$ at each turn $t$. The final policy objective is formulated based on Eq. ~(\ref{equ:module_level_policy_gradient}), and the clipped surrogate objective~\citep{schulman2017proximal}:
\begin{multline}
\label{equ:policy_loss}
\mathcal{L}_\pi(\phi) = -\mathbb{E} \Bigg[\sum_{t=0}^{T} \sum_{m=1}^M \min \bigg( \zeta_{(t,m)}\hat{A}_{(t,m)} ,\\
\text{clip}(\zeta_{(t,m)},1-\epsilon, 1+\epsilon) \hat{A}_{(t,m)} \bigg) \Bigg]
\end{multline}
$\zeta_{(t,m)} = \pi_\phi(\bm{y}_{(t,m)} | \bm{x}_{(t,m)}) / \pi_{\phi_\text{old}}(\bm{y}_{(t,m)} | \bm{x}_{(t,m)})$ is the ratio of probabilities between the policy being updated and the policy before update in each iteration.

Following previous studies~\citep{ziegler2019fine, NEURIPS2020_1f89885d}, to prevent the probability distribution output by the policy network $\pi_\phi^\text{RL}$ from deviating too far from the original $\pi_\phi^\text{IL}$ obtained by IL and breaking the output, we used a penalty based on Kullback–Leibler (KL) divergence as the final immediate reward $r_{(t,m)}$:
\begin{multline}
\label{equ:immediate_reward}
r_{(t,m)} = R(t,m) - \beta\mathbb{D}_\text{KL}\big[\pi_\phi^\text{RL}(\bm{y}_{(t,m)}|\bm{x}_{(t,m)}) \| \\
\pi_\phi^\text{IL}(\bm{y}_{(t,m)}|\bm{x}_{(t,m)})\big]
\end{multline}
where $\beta$ denotes a hyperparameter indicating the KL penalty coefficient. Algorithm~\ref{alg:ppo_algorithm} summarizes the RL algorithm.

\begin{algorithm}[ht]
\caption{UniPPN Optimization Algorithm with PPO}
\label{alg:ppo_algorithm}
\begin{algorithmic}[1] 
\REQUIRE Policy network $\pi_\phi$, value function $V_\psi$\\
\REQUIRE Dialogue system $\mathcal{F}$, dialogue environment $\mathcal{P}$
\STATE Install $\pi_\phi$ to the dialogue system $\mathcal{F}$\\
\FOR{each iteration}
\STATE Initialize replay buffer $\mathcal{D}$
\WHILE{$|\mathcal{D}| < \text{batch size } B \times M$}
\STATE // $\mathcal{F}$ interacts with $\mathcal{P}$
\FOR{each turn $t$}
\FOR{each $\text{Module}_m$}
\STATE Sample $\bm{y}_{(t,m)}$ $\sim$ $\pi_\phi(\bm{y}_{(t,m)} | \bm{x}_{(t,m)})$ by post-processing the output of $\text{Module}_m$ \\
\STATE Calculate reward $r_{(t,m)}$ by Eq.~(\ref{equ:immediate_reward})\\
\STATE Add $(\bm{x}_{(t,m)}, \bm{y}_{(t,m)}, r_{(t,m)})$ to $\mathcal{D}$
\ENDFOR
\ENDFOR
\ENDWHILE
\STATE Compute advantage estimate $\hat{A}_{(t,m)}$ by Eq.~(\ref{equ:advantage_estimate})
\FOR{each inner epoch}
\STATE Update $\psi$ by Eq.~(\ref{equ:value_loss})
\STATE Update $\phi$ by Eq.~(\ref{equ:policy_loss})
\ENDFOR
\ENDFOR
\end{algorithmic}
\end{algorithm}

\section{Details of UniPPN Training}
\label{appendix:sec:details_unippn_training}
We adopted the 355M parameter version of GPT-2 as the backbone model for UniPPN. The outputs of some modules, such as NLU and DST contain redundant occurrences of domain names or other symbols (e.g., \texttt{[restaurant, -, price, range, =, cheap, ; restaurant, -, food, =, italian]} with 12 tokens). This redundancy may complicate LM's generation task and hinder exploration during RL. To address this, we add tokens representing domain and slot name pairs and assign one token to a pair to simplify the sequence format (e.g., \texttt{[restaurant-pricerange=, cheap, restaurant-food=, italian]} with four tokens). This enables the model to focus solely on learning to modify $\text{out}_{(t,m)}$. The embedding parameters for these new vocabularies were randomly initialized. Below, we describe the learning details of IL and RL, as well as the impact of the module-level MDP.

\subsection{Imitation Learning}
In the supervised fine-tuning of $\pi_\theta$ using $D_{1:M}$, we divided $D_m$ into mini-batches of size 64 and trained them for 10 epochs using the AdamW optimizer. For the first five epochs, we trained only the embedding vectors of the newly added vocabulary for the slots with a learning rate of $5 \times 10^{-3}$, whereas for the latter five epochs, we trained the entire $\pi_\theta$ with $5 \times 10^{-5}$. This prevents random gradients caused by randomly initialized embedding parameters from destroying $\theta$ as a whole during the early stages of learning.

\subsection{Reinforcement Learning}
For $\pi$'s generation parameters during RL, we set the maximum number of input and output tokens to 256 and 128, respectively, and both the top $p$ and temperature to 1.0, to promote exploration. The total number of iterations for the entire learning process was set to 200, and the number of turns sampled in each iteration was 1,024. In each iteration, the trajectories of the post-processing outputs from all modules accumulated in the replay buffer were trained for four epochs with a mini-batch size of 128 and gradient accumulation steps $M$. We used the Adam Optimizer with a learning rate of $1 \times 10^{-6}$. $\gamma$ and $\lambda$ were set to 0.99 and 0.95 respectively, and the coefficient $\beta$ of the KL divergence penalty was set to 0.01.

\subsection{Impact of Module-level MDP}
\label{appendix:sec:impact_module_level_mdp}

We examined the contribution of the module-level MDP, introduced in our optimization algorithm, to UniPPN learning. We trained UniPPN for all the modules of SYS$_\text{PPO}$ with and without module-level MDP. During training without module-level MDP, all actions performed by UniPPN in the same turn received the same value estimation (called turn-level MDP). For each setup, the training was conducted using three random seeds.

\begin{figure}[tbp]
\begin{minipage}[t]{1\linewidth}
\centering
\includegraphics[width=\linewidth]{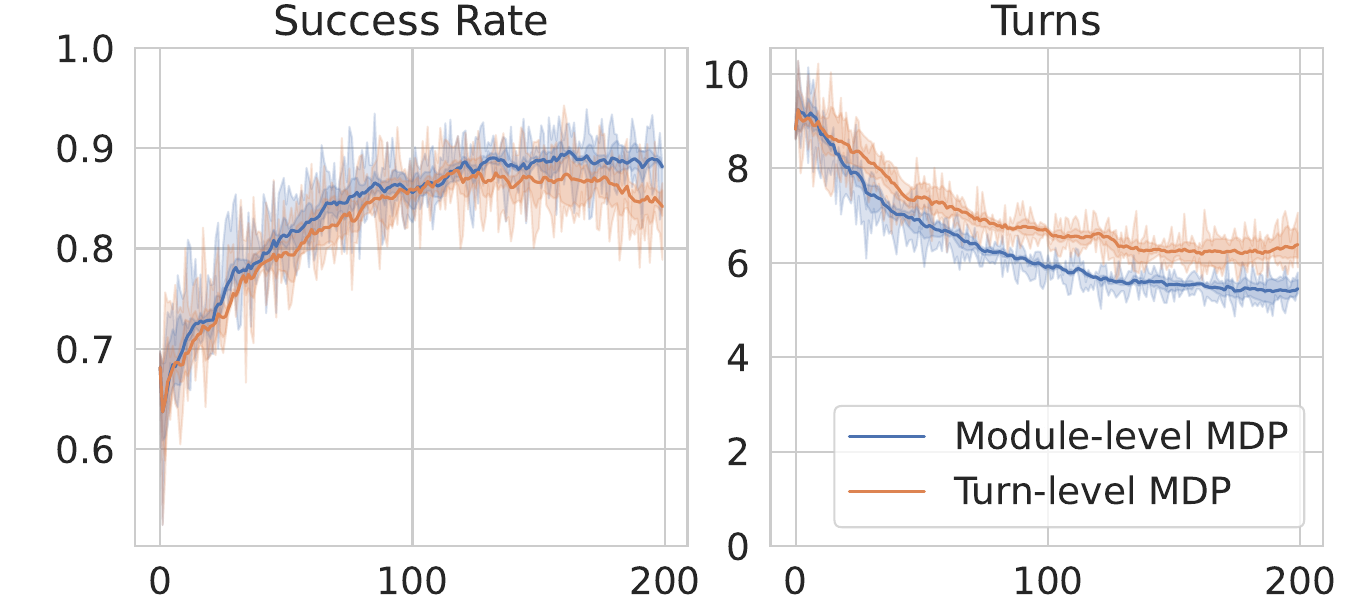}
\subcaption{Task completion metrics}
\label{fig:unippn_learning_curve_tasks}
\end{minipage}
\begin{minipage}[t]{1\linewidth}
\centering
\includegraphics[width=\linewidth]{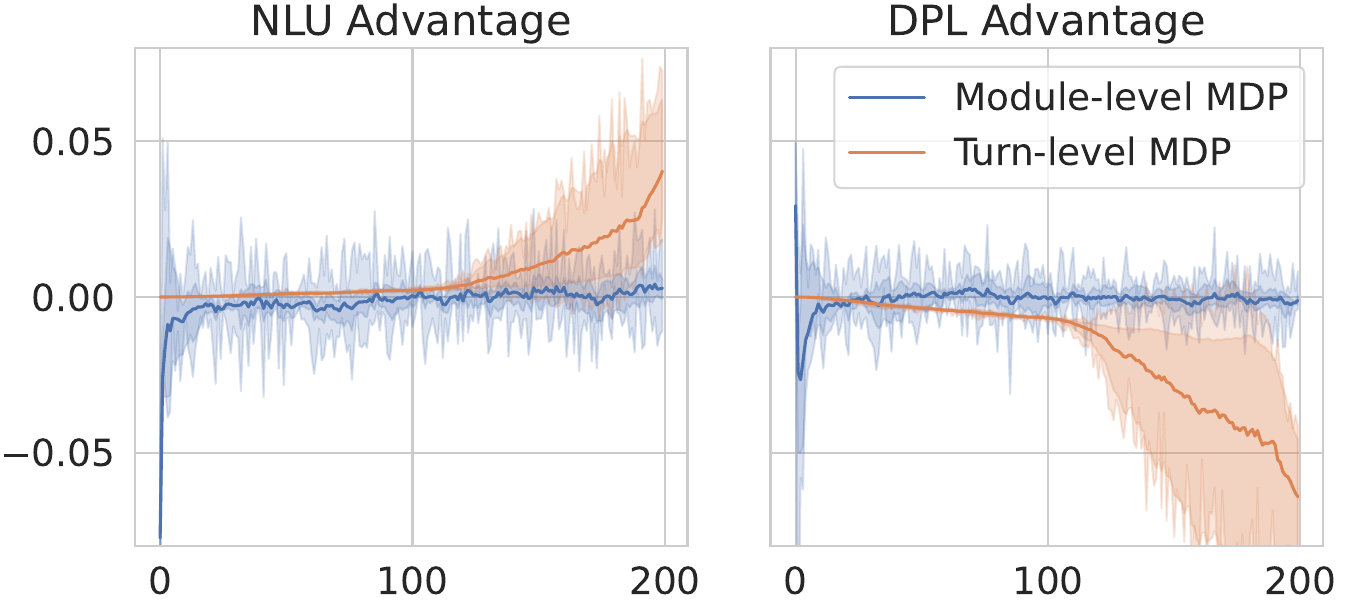}
\subcaption{Advantage estimates}
\label{fig:unippn_learning_curve_advantages}
\end{minipage} 
\caption{Learning curve where UniPPN was applied to SYS$_{\text{PPO}}$}
\label{fig:unippn_learning_curve}
\end{figure}

Figure~\ref{fig:unippn_learning_curve_tasks} shows the learning curves for the success rate and number of turns. The introduction of the module-level MDP improved the scores. Furthermore, turn-level MDP becomes particularly unstable in the latter half of learning, and the final performance decreases. To investigate this, we plotted the advantage estimates for the post-processing of each module (Figure~\ref{fig:unippn_learning_curve_advantages}). The plots reveal that advantage estimation is significantly destabilized and destroyed in the turn-level MDP. This can be considered an example of a credit assignment problem in RL. Specifically, assigning the same value estimation to all actions (i.e., the post-processing outputs of all modules) performed within the same turn fails to estimate the proper contribution of each action,  resulting in learning instability. This instability is believed to significantly affect the final system performance.

\section{Human Evaluation Details}
\label{appendix:sec:human_evaluation_detail}
The procedure of the human evaluation experiment followed these steps. First, each worker read a user goal randomly created for each dialogue. Next, the worker conducted a dialogue with one of the three systems (SYS$_\text{PPO}$, +BinPPN\&GenPPN, +UniPPN). The maximum number of turns for the dialogue was set to 20, which is the same as that used in the automatic evaluation experiment. Each worker judges whether a goal has been achieved within the maximum number of turns. After the dialogue ended, the participants answered three subjective evaluation questionnaires. To ensure the quality of the subjects, we only recruited workers with masters qualifications\footnote{\url{https://www.mturk.com/worker/help#what_is_master_worker}} from AMT. In addition, we restricted each worker from participating in the experiment only once. The reward for each participant was set to \$2, considering the time required per task (around ten minutes or less) and the minimum wage in the U.S.

\begin{table}[tbp]
\centering
\includegraphics[width=\linewidth]{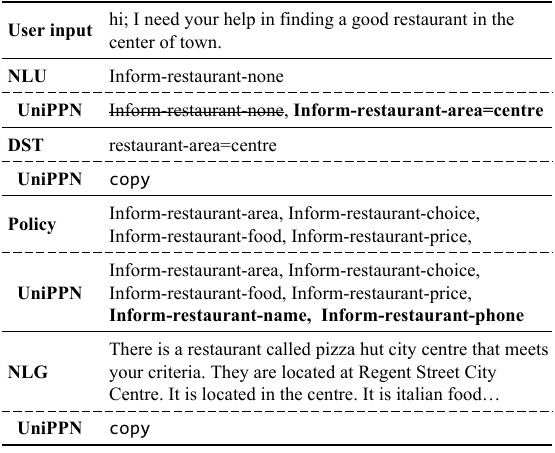}
\caption{Dialogue history between SYS$_\text{PPO}$ with UniPPN and a crowdworker from the human evaluation experiment. The rows below each module (that is, NLU, DST, policy, and NLG) display the results of post-processing by UniPPN. Text with strikethrough indicates information deleted by UniPPN, while bold text representsand information added.}
\label{tab:case_study}
\end{table}

To examine the performance of UniPPN, we quantitatively analyzed the dialogue history between the systems and crowdworkers. Table~\ref{tab:case_study} displays the response history for SYS$_\text{PPO}$ with UniPPN applied. Initially, for the user's input utterance, the original NLU failed to recognize that the area was ``centre'' and only output that the user was looking for a restaurant. UniPPN corrected the output by adding ``area=centre.'' For the DST output, UniPPN judged that it was not problematic and maintained (copied) the output without modifications. Regarding policy output, UniPPN added new DAs to restaurant names and phone numbers. This is probably because, during RL with the user simulator, UniPPN learned that adding a name and other information increased the likelihood of task success. Note that in dialogues with humans, outputting excessive information not explicitly requested by the user might negatively impact dialogue satisfaction scores. From the analysis, we confirmed cases where UniPPN, developed in a simulation environment, was also effective in dialogue with humans, resulting in improved task completion performance.

\fi

\end{document}